\documentclass{article}

\usepackage{graphicx} % more modern
\usepackage{subfigure}
\usepackage{amssymb,amsmath,color}

\usepackage{natbib}
\usepackage{algorithm}
\usepackage{algorithmic}

\usepackage{hyperref}

\usepackage[accepted]{icml2012}
\newcommand{\mysec}[1]{Section~\ref{sec:#1}}
\newcommand{\eq}[1]{Eq.~(\ref{eq:#1})}
\newcommand{\myfig}[1]{Figure~\ref{fig:#1}}
    
\def \E{{\mathbb E}}

\def \F {\mathcal{F} }
\def \Ft { \widetilde{\mathcal{F}} }
\def \X { \mathcal{X}  }
\def \A { \mathcal{A} }

\newcommand{\rb}{\mathbb{R}}
\newcommand{\BEAS}{\begin{eqnarray*}}
\newcommand{\EEAS}{\end{eqnarray*}}
\newcommand{\BEA}{\begin{eqnarray}}
\newcommand{\EEA}{\end{eqnarray}}
\newcommand{\BEQ}{\begin{equation}}
\newcommand{\EEQ}{\end{equation}}
\newcommand{\BIT}{\begin{itemize}}
\newcommand{\EIT}{\end{itemize}}
\newcommand{\BNUM}{\begin{enumerate}}
\newcommand{\ENUM}{\end{enumerate}}
\newcommand{\BA}{\begin{array}}
\newcommand{\EA}{\end{array}}

\newenvironment{proofsketch}{\par\noindent{\bf Proof sketch.~}}{\hfill\BlackBox\\[2mm]}

\newtheorem{proposition}{Proposition}

\newcommand{\BlackBox}{\rule{1.5ex}{1.5ex}}  % end of proof

\icmltitlerunning{On the Equivalence between Herding and Conditional Gradient Algorithms}

\begin{document}

\twocolumn[
\icmltitle{On the Equivalence between Herding \\and Conditional Gradient Algorithms}

% unless you've provided the [accepted] option to the icml2012
\icmlauthor{Francis Bach, Simon Lacoste-Julien, Guillaume Obozinski}{firstname.lastname@inria.fr}
\icmladdress{Sierra project-team, INRIA, D\'epartement d'Informatique de l'Ecole Normale Sup\'erieure, Paris, France}

\icmlkeywords{boring formatting information, machine learning, ICML}

\vskip 0.3in
]

\begin{abstract}

%\vspace*{-.04cm}

We show that the herding procedure of~\citet{welling2009herding} takes exactly the form of a standard convex optimization algorithm---namely a conditional gradient algorithm minimizing a quadratic moment discrepancy. This link enables us to invoke convergence results from convex optimization and to consider faster alternatives for the task of approximating integrals in a reproducing kernel Hilbert space. We study the behavior of the different variants through numerical simulations.
Our experiments shed more light on the learning bias of  herding: they indicate that
while we can improve over herding on the task of approximating
integrals, the original herding algorithm
approaches more often the maximum
entropy distribution.

\vspace*{-.1cm}

\end{abstract}

\vspace*{-.1cm}

\section{Introduction}

\vspace*{-.05cm}

The herding algorithm has recently been presented by~\citet{welling2009herding} as a computationally attractive alternative method for learning in intractable Markov random fields models (MRF). Instead of first estimating the parameters of the MRF by maximum likelihood / maximum entropy (which requires approximate inference to estimate the gradient of the partition function), and then sampling from the learned MRF to answer queries, herding directly generates \emph{pseudo-samples} in a deterministic fashion with the property of asymptotically matching the empirical moments of the data (akin to maximum entropy). The herding algorithm generates pseudo-samples $x_t$ with the following simple recursion:

\vspace*{-.45cm}

\begin{equation} \label{eq:herding_welling}
\begin{aligned}
x_{t+1} &  \in \arg \max_{ x \in \X } \ \langle w_t, \Phi(x) \rangle \\[-.1cm]
w_{t+1} & = w_t + \mu - \Phi(x_{t+1}),
\end{aligned}
\end{equation}

\vspace*{-.35cm}

where $\X$ is the observation space; $\Phi$ is a feature map from $\X$ to $\F$, which could be viewed as the vector of sufficient statistics for some exponential family, and $\mu$ is a mean vector to match (the empirical moment vector of the same family). Unlike in frequentist learning of MRFs, the parameter $w_t$ never converges to a point in herding and actually follows a ``weakly chaotic'' walk~\citep{welling2010statistical}.

The herding updates can be motivated from two different perspectives. From the \emph{learning perspective}, the herding updates can be derived by performing fixed-step-size subgradient ascent on the zero-temperature limit of the annealed likelihood function of the MRF---called the ``tipi function'' by~\citet{welling2009herding}. From this perspective,   herding   was later generalized to MRFs with latent variables~\citep{welling2009UAIherding} as well as discriminative MRFs~\citep{gelfand2010herding}.

From the \emph{moment matching} perspective, which has been explored more in details by~\citet{chensuper}, the herding updates can be derived as an effective way to choose \emph{greedily} pseudo-samples $x_t$ in order to quickly decrease the moment discrepancy $\mathcal{E}_t \doteq \| \mu - \frac{1}{t} \sum_{i=1}^t \Phi(x_i) \|$~\citep{chensuper}. Under suitable regularity conditions, $\mathcal{E}_t$ decreases as $O(1/t)$ for the herding updates---this is faster than i.i.d.~sampling from the distribution generating $\mu$ (e.g., the training data) which would yield the slower $O(1/\sqrt{t})$ rate. This faster rate has been explained by \emph{negative} auto-correlations amongst the pseudo-samples and was used by~\citet{chensuper} to sub-select a small collection of representative ``super-samples'' from a much larger set of i.i.d.~samples. %This fast moment matching perspective is the focus of this paper.
We make the following contributions:

\vspace*{-.4cm}

\begin{list}{\labelitemi}{\leftmargin=1.1em}
   \addtolength{\itemsep}{-.6\baselineskip}

\item[--] We show that herding as described by \eq{herding_welling} is equivalent to a specific type of conditional gradient algorithm (a.k.a.~Frank-Wolfe algorithm) for the problem of estimating the mean~$\mu$. This provides a novel understanding and another explicit cost function that herding is minimizing.

\item[--] This interpretation yields improvements, \emph{for the task of estimating means}, with other faster variants of the conditional gradient algorithm, which lead to \emph{non-uniform weights}, one based on line-search, one based on an active-set algorithm.

\item[--] Based on existing results from convex optimization, we extend and improve the convergence results of herding. In particular, we provide a linear convergence rate for the line-search variant in finite-dimensional settings and show how the conditions assumed by~\citet{chensuper} in fact never hold in the infinite-dimensional setting.

\item[--] We run experiments that show that algorithms which estimate faster the mean than herding
generate samples that are not better (and typically worse) than the ones obtained with herding when evaluated in terms of the ability to approximate a sample with large entropy, a property which (if or when satisfied by herding) could be the basis for an interpretation of herding as a learning algorithm~\citep{welling2009herding}.
%These results could help shedding more light on what is (and is not) the learning bias of herding.

\end{list}

\vspace*{-.35cm}

\section{Mean estimation}

\vspace*{-.1cm}

\label{sec:setup}
We start with a similar setup as~\citet{chensuper}, where herding can be interpreted as a way to approximate integrals of functions in a reproducing kernel Hilbert space (RKHS). We consider a set $\X $ and a mapping $\Phi$ from $\X $ to a RKHS $\F $. Through this mapping, all elements of $\F $ may be identified with real functions~$f$ on $\X $ defined by $f(x) = \langle f, \Phi(x) \rangle$, for $x \in \X $. We denote by $k:(x,y) \mapsto k(x,y)$ the associated positive definite kernel. Note that the mapping $\Phi$ may be explicit (classically in low-dimensional settings) or implicit---where the kernel trick can be used,~see \mysec{comp} and \citet{chensuper}.

Throughout the paper, we assume that the data is uniformly bounded in feature space, that is,  for all $x \in \X $, $\| \Phi(x) \| \leqslant R$, for some $R>0$; this condition is needed for the updates of \eq{herding_welling} to be well-defined.

We denote by $\mathcal{M} \subset \F $ the \emph{marginal polytope} \citep{wainwright2008graphical,chensuper}, i.e., the convex-hull of all vectors $\Phi(x)$ for $x \in \X $. Note that for any $f \in \F $, we have

\vspace*{-.45cm}

$$
\sup_{x \in \X } f(x)  = \sup_{ g \in \mathcal{M} } \langle f, g \rangle ,
$$

\vspace*{-.25cm}

and that $|f(x)| = | \langle f, \Phi(x) \rangle | \leqslant \|f\| R$ for all $x \in \X $ and $f \in \F $ (i.e., all functions with finite norm are bounded).

\label{sec:kt}

\begin{figure}

 \vspace*{-.25cm}

\centering
\includegraphics[scale=.5]{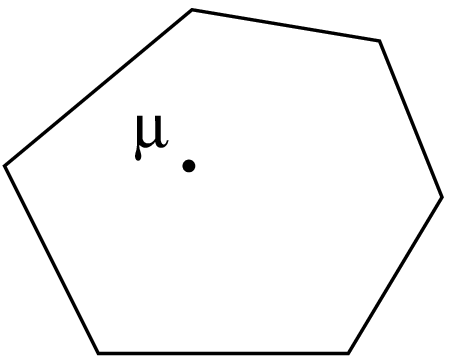} \hspace*{1cm}
\includegraphics[scale=.5]{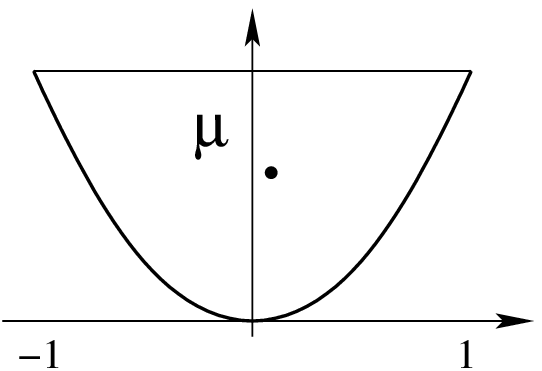}

\vspace*{-.25cm}

 \caption{Marginal polytope in two situations: (left) finite number of extreme points (typical with discrete data),   (right) polynomial kernel in one dimension, with $\Phi(x)=(x,x^2)$ for $x \in [-1,1]$.}
 \label{fig:illustration}
 \vspace*{-10pt}
\end{figure}

\textbf{Extreme points of the marginal polytope.} \hspace*{.1cm}
In all the cases we consider in \mysec{simulations},  it turns out that all points of the form $\Phi(x)$, $x \in \X $ are extreme points of $\mathcal{M}$ (see an illustration in \myfig{illustration}).
This is indeed always true when $\|\Phi(x)\|$ is constant for all $x \in \X $
(for example for our infinite-dimensional kernels on $[0,1]$ in \mysec{01}); it is also true if $\Phi(x)$ contains both an injective feature map $\widetilde{\Phi}(x)$ and its self-tensor-product  $\widetilde{\Phi}(x) \otimes \widetilde{\Phi}(x)$, which is the case in the graphical model examples in \mysec{gm}.

\textbf{Mean element and expectation.} \hspace*{.1cm}
We consider a fixed probability distribution $p(x)$ over~$\X $. Following~\citet{chensuper}, we denote by $\mu$ the mean element~\citep[see, e.g.,][]{smola2007hilbert}:

\vspace*{-.45cm}

$$\mu = \E_{p(x)} \Phi(x) \in \mathcal{M} .$$

\vspace*{-.25cm}

Note that in the learning perspective, $p$ is the empirical distribution on the data and so $\mu$ is the corresponding empirical moment vector to match. To approximate this mean, we consider $n$ points $x_1,\dots,x_n \in \X $ combined linearly with positive weights $w_1,\dots,w_n$ that sum to one. These define $\hat{p}$, the associated weighted empirical distribution, and $\hat{\mu}$ the approximating mean:

\vspace*{-.45cm}

\begin{equation} \label{eq:weighted_mean}
\textstyle
\hat{\mu} = \E_{\hat{p}(x)} \Phi(x) = \sum_{i=1}^n w_i \Phi(x_i) \in \mathcal{M} .
\end{equation}

\vspace*{-.25cm}

For all functions $f \in \F $,  we then have

\vspace*{-.5cm}

$$
\E_{p(x)} f(x) = \E_{p(x)} \langle f, \Phi(x) \rangle = \langle \mu, f \rangle,
$$

\vspace*{-.2cm}

and similarly $\E_{\hat{p}(x)} f(x) = \langle \hat{\mu}, f \rangle$.
We thus get, using Cauchy-Schwarz inequality,

\vspace*{-.55cm}

$$ \textstyle
\sup_{f \in \F , \ \|f\|=1} |\E_{p(x)} f(x) - \E_{\hat{p}(x)} f(x) | = \| \mu - \hat{\mu} \| ,
$$

\vspace*{-.25cm}

and controlling $\mu - \hat{\mu}$ is enough to control the error in computing the expectation for all $f \in \F $ with finite norm. Note that a random i.i.d.~sample from $p(x)$ would lead to an expected worst-case error which is less than $\frac{4 R }{ \sqrt{n}}$---a classical result based on Rademacher averages~\citep[see, e.g.][]{boucheron2005theory}.

In this paper, we will try to find a good estimate $\hat{\mu}$ of $\mu$ based on a \emph{weighted} set of points from $\{\Phi(x), x \in \X \}$, generalizing~\citet{chensuper}, and show how this relates to herding.

%\vspace*{-.3cm}

\vspace*{-.1cm}

\section{Related work}

\vspace*{-.1cm}

This paper brings together three lines of work, namely the approximation of integrals, herding and convex optimization. The links between the first two were clearly outlined by~\citet{chensuper}, while the present paper provides the novel interpretation of herding as a well-established convex optimization algorithm.

\subsection{Quadrature/cubature formulas}

\vspace*{-.1cm}

The evaluation of expectation, or equivalently of integrals, is a classical problem in numerical analysis. When the input space $\mathcal{X}$ is a compact subset of $\rb^p$ and $p(x)$ is the density of the distribution with respect to the Lebesgue measure, then this is equivalent to evaluating the  integral $\int_{\mathcal{X}} f(x) p(x) dx$. Quadrature formulas are aimed at computing such integrals as a weighted combinations of values of~$f$ at certain points, which is exactly the problem we consider in \mysec{setup}.

Although a thorough review of quadrature formulas is outside of the scope of this paper, we mention two methods which are related to our approach.
 First, given a positive definite kernel and a given set of points (typically sampled i.i.d.~from a given distribution), the Bayes-Hermite quadrature of~\citet{o1991bayes} essentially computes an orthogonal projection of $\mu$ onto the affine hull of this set of points. This does not lead to positive quadrature weights, and one may thus replace the affine hull by the convex hull to obtain such nonnegative weights, which we do in our experiments in \mysec{experiments}.

Moreover, quasi-Monte Carlo methods consider sequences of so-called ``quasi-random''  quadrature points so that the empirical average approaches the integral. These quasi-random sequences are such that the approximation error goes down as $O(1/n)$ (up to logarithmic terms) for functions of bounded variation, as opposed to $O(1/\sqrt{n})$ for a random sequence. In simulations, we used a Sobol sequence~\citep[see, e.g.,][]{morokoff1994quasi}.

\subsection{Franke-Wolfe algorithms}

\vspace*{-.1cm}

\begin{figure}

 \vspace*{-.25cm}

\centering
\includegraphics[scale=.5]{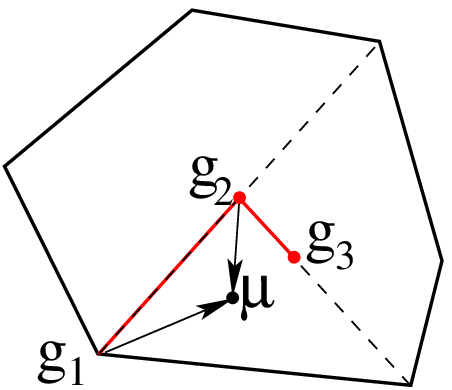}  \hspace*{1cm}
\includegraphics[scale=.5]{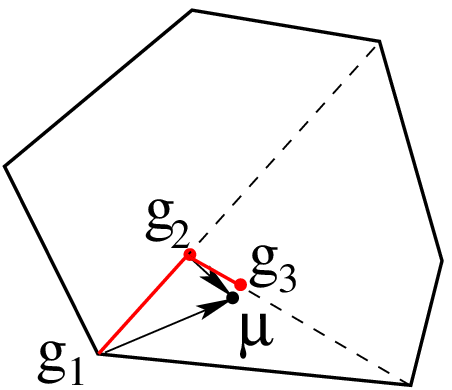}

\vspace*{-.25cm}

 \caption{Two versions of two iterations of conditional gradient after moving to an initial corner $g_1$; left: $\rho_t = \frac{1}{t+1}$, right: line search. The minimum-norm-point algorithm would have converged to $\mu$ after two iterations.}
 \label{fig:illustrationcg2}
 \vspace*{-10pt}
\end{figure}

Given a smooth (twice continuously differentiable) \emph{convex} function $J$ on a \emph{compact convex set} $\mathcal{M}$ with gradient $J'$, Frank-Wolfe algorithms are a class of iterative optimization algorithms that only
require (in addition to the computation of the gradient $J'$) to be able to optimize linear functions on $\mathcal{M}$. The first class of such algorithms is often referred to as \emph{conditional gradient} algorithms: given an iterate $g_t$, the minimum of $\langle J'(g_t), g \rangle$ over $g \in \mathcal{M}$ is computed, and the next iterate is taken on the segment between $g_t$ and $g$, i.e., $g_{t+1} = \rho_t g_t + ( 1 -\rho_t) g$, where $\rho_t \in [0,1]$. There are two natural choices for $\rho_t$, (a) simply taking $\rho_t = 1/(t+1)$ and (b) performing a line search to find the point in the segment with optimal value (either for $J$ or for a quadratic upper-bound of~$J$). These are illustrated in \myfig{illustrationcg2}, and convergence rates are detailed in \mysec{rates}. Moreover, for quadratic functions, the conditional gradient algorithm with step sizes $\rho_t = 1/(t+1)$ has a dual interpretation as subgradient ascent~\citep[see, e.g.,][]{bach2011learning}, which we outline in \mysec{subgrad}.

Finally, in order to minimize the number of iterations, a variant known as the \emph{minimum-norm-point} algorithm will find $g_{t+1}$ that minimizes $J$ on the  convex hull of \emph{all} previously visited points, using a specific active-set algorithm~\citep[see][Sec.~6, for details]{bach2011learning}. For convex sets with finitely many extreme points, it converges in a finite number of iterations with higher (though still polynomial) iteration computational cost~\citep{wolfe1976finding}.

%\begin{figure}
%% \includegraphics[scale=.5]{illustration.eps}
%\centering
%\includegraphics[scale=.8]{cg.eps}  \hspace*{.21cm}
%\includegraphics[scale=.8]{cg_ls.eps}
% \caption{Two versions of one iteration of conditional gradient; left: $\rho_t = \frac{1}{t+1}$, right: line search.}
% \label{fig:illustrationcg}
%\end{figure}

%\begin{figure}
%% \includegraphics[scale=.5]{illustration.eps}
%\centering
%\includegraphics[scale=.8]{polytope_algo.eps}
% \caption{Herding with $\rho_t = \frac{1}{t+1}$.}
% \label{fig:illustrationherding}
%\end{figure}
%

\vspace*{-.1cm}

\section{Herding as a Frank-Wolfe algorithm}

\vspace*{-.1cm}

To relate herding with   conditional gradient algorithms, we consider the following optimization problem:

\vspace*{-.45cm}

\BEQ
\label{eq:herding}
\min_{ g \in \mathcal{M} } J(g) = \frac{1}{2} \| g - \mu \|^2,
\EEQ

\vspace*{-.35cm}

with the trivial unique solution $\mu$.
A conditional gradient algorithm to solve this optimization problem with stepsize $\rho_t = 1/(t+1)$ use the iterates

\vspace*{-.8cm}

\BEA
\label{eq:herding_cg}
\nonumber \bar{g}_{t+1} & \in &  \arg \min_{ g \in \mathcal{M}} \, \langle  g_t - \mu,g \rangle , \\[-.1cm]
g_{t+1}  & = &  (1 - \rho_{t})\, g_t + \rho_{t} \, \bar{g}_{t+1}.
\EEA

\vspace*{-.4cm}

But these updates are exactly the same as herding via the change of variable $g_t = \mu - w_t/t$.
Indeed,  the minimizer of a linear function of a convex set $\bar{g}_{t+1}$ can be restricted to be an extreme point of $\mathcal{M}$; this implies that $\bar{g}_{t+1}  = \Phi(x_{t+1})$ for a certain $x_{t+1}$. The herding updates in \eq{herding_welling} are thus equivalent to the conditional gradient minimization of $J$ with step size given by
$\rho_t = 1/(t+1)$.

With this choice of step size, we get
$(t+1) g_{t+1} = t g_t + \Phi(x_{t+1})$, that is
$
g_t = \frac{1}{t}\sum_{u=1}^{t} \Phi(x_u),
$
and we thus get uniform weights in Eq.~\eqref{eq:weighted_mean}.

For general step-sizes $\rho_t \in [0,1]$, if we assume that we start the algorithm with $\rho_0 = 1$ (which implies $g_1 = \Phi(x_1)$), then we get that
$
g_t = \sum_{u=1}^t \big( \prod_{v = u}^{t-1} ( 1 - \rho_{v-1}) \rho_{u-1} \big) \Phi(x_u),
$
which thus leads to non-uniform weights in Eq.~\eqref{eq:weighted_mean}, though they still sum to one. The conditional gradient updates in Eq.~\eqref{eq:herding_cg} can thus be seen as a generic algorithm to obtain a weighted set of points to approximate $\mu$, and traditional herding is the $\rho_t = 1/(t+1)$ step-size case.

A second choice of step-size for $t \geq 1$ is to use a line search, which leads in this setting (where the global unconstrained minimum happens to belong to $\mathcal{M}$) to
$
\rho_{t}  = \frac{\langle g_t - \mu , g_t - \bar{g}_{t+1} \rangle }{\| \bar{g}_{t+1} - g_{t} \|^2} \in [0,1].
$
This leads to a variant of herding with non-uniform weights.

We finally comment on the initialization $g_0$ for the updates in Eq.~\eqref{eq:herding_cg}. In the kernel herding algorithm of~\citet{chensuper}, the authors use $w_0 = \mu$ as this is required to interpret the herding updates as greedily minimizing $\mathcal{E}_t$ (with the additional assumptions that $\| \Phi(x) \|$ is constant). In our setting, this corresponds to choosing $g_0 = 0$ (which might be outside of  $\mathcal{M}$, though this is not problematic in practice). Another standard choice (for MRFs in particular) is to use $w_0 = 0$ ($g_0 = \mu$), which means that the first point $x_1$ is chosen randomly from the extreme points of $\mathcal{M}$---this is the scheme we used. As is common in convex optimization, we didn't see any qualitative difference in our experiments between the two types of initialization.

\subsection{Dual problem and subgradient descent}

\vspace*{-.1cm}

\label{sec:subgrad}
\citet{welling2009herding} proposed originally an algorithmic interpretation of herding as performing subgradient ascent with constant step size on a function obtained as the zero temperature limit of the log-likelihood of
an exponential model that he called the ``tipi function". Our interpretation of the procedure as a Frank-Wolfe algorithm might therefore appear as a conflicting interpretation at first sight. To establish a natural connection between these two interpretations, we can compute the Fenchel-dual optimization problem to \eq{herding}. Indeed, we have (with standard arguments for swapping the min and max operations):

\vspace*{-.75cm}

\BEAS
 \!\!\min_{ g \in \mathcal{M}} \frac{1}{2} \|  g - \mu \|^2
& \!\!=\!\! &  \min_{ g \in \mathcal{M}} \max_{f \in \F } f^\top ( g - \mu) -\frac{1}{2} \| f\|^2 \\[-.1cm]
& \!\! =\!\!  & \max_{f \in \F }   \min_{ g \in \mathcal{M}}  f^\top ( g - \mu) -\frac{1}{2} \| f\|^2 \\[-.1cm]
& \!\! =\!\!  & \max_{f \in \F }  \big\{  \min_{  x \in \X  }  f(x) - \langle f, \mu \rangle -\frac{1}{2} \| f\|^2 \big\}.
\EEAS

\vspace*{-.35cm}

The dual function $f \mapsto  \min_{  x \in \X  }  f(x) - \langle f, \mu \rangle -\frac{1}{2} \| f\|^2$ is $1$-strongly concave and non-differentiable, and a natural algorithm to maximize it is thus subgradient ascent with a step size equal to $\frac{1}{t+1}$, which is known to be equivalent to the primal conditional gradient algorithm with step sizes $\rho_t = 1/(t+1)$~\citep[][App.~A]{bach2011learning}. It is therefore not surprising that
herding is equivalent to subgradient ascent with a decreasing stepsize on this function  (with the identification $f_t= g_t- \mu = - w_t/t$). The presence of the squared norm which is added to the ``tipi function" merely reflects the change of scaling between $g_t$ and $w_t$. It is worthwhile mentioning that other functions, like Bregman divergences, would have led to a different dual function hence adding a different term than a squared norm  to the ``tipi function".

\subsection{Convergence analysis}

\vspace*{-.1cm}

\label{sec:rates}
Without further assumptions on the problem, then the two choices of step sizes
lead to a convergence rate of the form~\citep{dunn1980convergence,bach2011learning}:

\vspace*{-.35cm}

$$\frac{1}{2} \| g_t - \mu\|^2 \leqslant 4 \frac{R^2}{t},$$

\vspace*{-.25cm}

where $R$ is diameter of the marginal polytope.
Note that the convergence in $O(1/t)$ does not lead to improved estimation of integrals over random sampling. Moreover, without further assumptions, current theoretical analysis does not allow to distinguish between the two forms of conditional gradient algorithms (although they differ a bit in practice, see \mysec{simulations}).

However, if we assume that within the affine hull of $\mathcal{M}$, there exists a ball of center $\mu$ and radius $d>0$ that is included in $\mathcal{M}$ (i.e., $\mu$ is in the relative interior of $\mathcal{M}$), then both choices of step sizes yield faster rates than random sampling. For the version with line search, we actually obtain a linear convergence rate~\citep{beck2004conditional}:

\vspace*{-.45cm}

$$\frac{1}{2} \| g_t - \mu\|^2 \leqslant R^2 \exp \big( - \frac{d^2 t }{R^2 } \big).$$

\vspace*{-.25cm}

For the version without line search ($\rho_t = 1/(t+1)$), \citet{chensuper} shows the slower convergence rate in $O(1/t^2)$:

\vspace*{-.75cm}

$$\frac{1}{2} \| g_t - \mu\|^2 \leqslant \frac{2 R^4}{d^2 t^2}.$$

\vspace*{-.35cm}

Concerning the assumption that $\mu$ is in the relative interior of $\mathcal{M}$, we now show that in finite-dimensional settings, this assumption is always satisfied under reasonable conditions, while it is \emph{never} satisfied in a large class of infinite-dimensional settings (namely for Mercer kernels).

We first provide an equivalent definition of this condition which is used in the proofs. Let $\A$ be the affine hull of $\mathcal{M}$, $\mu_0$ the orthogonal projection of $0$ on $\A$, and $\Ft$ the space of directions (or difference space) of $\A$ (i.e., $\Ft = \A - \mu_0$).\footnote{It turns out that $\mu_0$ is a function taking a constant value since the orthogonality condition yields $\langle \mu_0, \Phi(x) - \Phi(y) \rangle = 0$, i.e., $\mu_0(x) = \mu_0(y)$ for all $x,y \in \X$, by the reproducing property of $\F$.} Now there exists $d>0$ so that any element of $\A$ which is at distance less than $d$ of $\mu$ is in $\mathcal{M}$
if and only if $\forall f \in \F$, the maximum of $f^\top g$ over $g \in \A$ and $\| g - \mu  \|   \leqslant d$ is less than the maximum of $f^\top g$ over $g \in \mathcal{M}$. Given the properties of $\A$ and $\Ft$, this is equivalent to:

\vspace*{-.85cm}

\BEA
\nonumber
&&\forall f \in \Ft, \ \max_{ \| g - \mu  \|   \leqslant d   } f^\top g \leqslant \max_{g \in \mathcal{M}}
f^\top g
\\[-.1cm]
%\nonumber
&\Leftrightarrow & \forall f \in \Ft, \ \langle \mu, f \rangle + d \| f\|  \leqslant \max_{x \in \X  } f(x).
\label{eq:cond}
%&\Leftrightarrow &  \forall f \in \Ft, \
%    d \| f\|  \leqslant \max_{x \in \X  } f(x) -  \langle \mu, f \rangle  .
\EEA

\vspace*{-.5cm}

\begin{proposition}
Assume that $\F $ is finite-dimensional, that $\X $ is a compact topological measurable space with a continuous kernel function, and that the distribution $p$ has full support on~$\X $. Then $\exists d>0$ so that \eq{cond} is true.
\end{proposition}

\vspace*{-.3cm}

\begin{proofsketch}  \hspace*{-.5cm}
It is sufficient to show that $\Omega: f \mapsto \max_{x \in \X  } f(x) -  \langle \mu, f \rangle $ is a norm on $\Ft$: as all norms are equivalent in finite dimensions, we get $d \| f\| \leq \Omega(f)$ for some $d>0$, yielding \eq{cond}. $\Omega$ is convex and positively homogeneous by construction. Now $\Omega(f)=0$ implies that $\E_{p(x)}[ f(x) - \max_{y} f(y)] = 0$, and thus $f(x) = \max_{y} f(y)$ for $x$ in the support of $p$ (assumed to be $\X$) using the fact that $f$ is continuous (since the kernel is continuous), and so $f$ is a constant function. We then have two possibilities: either $\mu_0 = 0$, in which case one can show that there is no non-zero constant functions in $\F$; otherwise $f = \alpha \mu_0$ for some $\alpha$ and thus the orthogonality condition $\langle f, \mu_0 \rangle = 0$ implies that $\alpha=0$. Both cases imply $f=0$, hence $\Omega$ is a norm.
\end{proofsketch}

\vspace*{-.7cm}

\begin{proposition}
 Assume $\X $ is a compact subspace of $\rb^d$, and that the kernel $k$ is a continuous function on $\X  \times \X $. If $\F $ is infinite-dimensional, then there exists no $d>0$ so that \eq{cond} is true.
\end{proposition}

\vspace*{-.3cm}

\begin{proofsketch} \hspace*{-.1cm}
We can apply Mercer theorem to the kernel $\tilde{k}(x,y)$ of the projection onto the orthogonal of $\{ \mu, \mu_0\}$. This kernel is also a Mercer kernel, and we get
an orthonormal basis $(e_k)_{k \geqslant 1}$ of $L^2(\X )$ with \emph{countably many} eigenvalues $\lambda_k$ that are summable.  Moreover, $(\lambda_k^{1/2} e_k)_{k \geqslant 1}$ is known to be an orthonormal basis of the associated feature space $\mathcal{F}$~\citep{smale}, and for all $x,y \in \X$, $\tilde{k}(x,y) = \sum_{k \geqslant 1} \lambda_k e_k(x) e_k(y)$, with uniform convergence. This implies that for  $f_k = \lambda_k^{1/2} e_k$, we have $\| f_k\| = 1$, and $\langle f_k , \mu_0 \rangle = \langle f_k , \mu \rangle  = 0$.

If we  assume that there exists $d>0$ so that \eq{cond} is true, then we have for all $k \geqslant 1$,
 $\max_{x \in \X } |f_k(x)| \geqslant d  $. Since $\mathcal{X}$ is compact, there exists a cover of $\mathcal{F}$ with finitely many balls of radius $d/4R$. Let $\mathcal{Y}$ be the finite set of centers.
 Since all functions $f_k$ are Lipschitz-continuous with constant $2R$, then for all $k \geqslant 1$,
 $\max_{x \in \mathcal{Y} } |f_k(x)| \geqslant d  - 2R \times d/4R = d/2$. Since $\mathcal{Y}$ is finite, there exists $x \in \mathcal{Y}$ so that $ |f_k(x)| \geqslant d/2>0$ for infinitely many values of $k$; this contradicts the summability of  $\sum_{k \geqslant 1} f_k(x)^2$. Hence the result.
\end{proofsketch}

\vspace*{-.6cm}

The last proposition shows that the current theory only supports the slower rates of $O(1/t)$ for the two conditional gradient algorithms in infinite-dimensional settings. On the other hand, we note that, in some cases, traditional herding performs empirically better without known theoretical justification (see \mysec{simulations}).

\vspace*{-.2cm}

  \subsection{Computational issues}
  \label{sec:comp}

\vspace*{-.1cm}

 In order to run a herding algorithm, there are two potential computational bottlenecks:

 \textbf{Computing expectations $\langle \mu, \Phi(x) \rangle$}: in a learning context (empirical moment matching), these are done through an empirical average. In an integral evaluation context, in finite-dimensional settings, one   needs to compute $\E_{p(x)} \Phi(x)$; while in an infinite-dimensional setting, following~\citet{chensuper}, expectations of the form $\E_{p(x)}k(x,y)$ need to be computed. This can be done for some pairs of kernels/distributions, like the ones we choose in \mysec{experiments}, but not in general.

  \textbf{Minimizing $\langle g_t - \mu , \Phi(x) \rangle$ with respect to $x \in \mathcal{X}$}: in general, this computation may be relatively hard (it is for example NP-hard in the context of the graphical models we consider in \mysec{simulations}). In practice, \citet{chensuper} and \citet{welling2009UAIherding} perform local search, while another possibility is to perform the minimization through exhaustive search in a finite sample. Note that a convex relaxation through variational methods~\citep{wainwright2008graphical} could provide an interesting alternative.

%SLJ: TO INCLUDE IN REBUTTAL -> approximate maximization aspect...

\vspace*{-.1cm}

 \section{Experiments}

\vspace*{-.1cm}

 \label{sec:simulations}
 \label{sec:experiments}
The goals of these simulations are (a) to compare the different algorithms aimed at estimating integrals, i.e., assess herding for the task of mean estimation (\mysec{01} and \mysec{gm}), and (b) to briefly study the empirical relationship with maximum entropy estimation in a learning context (\mysec{bits}).

\vspace*{-.15cm}

\subsection{Kernel herding on $\X =[0,1]$}

\vspace*{-.1cm}

 \label{sec:01}
 \textbf{Problem set-up.} \hspace*{.1cm}
 In this section, we consider $\X  = [0,1]$ and the norm
 $\| f\|^2 = \int_0^1 [  f^{({\nu})}(x) ] ^2 dx$ on the infinite-dimensional space of functions with zero mean and whose $\nu$-th derivative exists and is in $L^2([0,1])$.
 As shown by~\citet{Wah:1990},
 the associated kernel is equal to $ \frac{B_{2 \nu}(x-y - \lfloor x - y\rfloor)}{(2\nu)!}$,
 where
 $B_{2\nu}$ is the $(2\nu)$-th Bernoulli polynomial, with
 $B_2(x) =   \textstyle x^2 - x + \frac{1}{6} $ and $
  B_6(x) =  \textstyle x^6 - 3x^5 + \frac{5}{2}x^4  - \frac{1}{2} x^2 + \frac{1}{42}$.

  We consider  either the uniform density on $[0,1]$, or a  randomly selected density of the form $p(x) \propto \big( \sum_{i=1}^d a_i \cos (2 i \pi x) + b_i \sin (2i \pi x) \big)^2$, for which all required expectations may be computed in closed form. In particular, the mean element is computed as $\mu: x \mapsto \E[k(Y,x)]$ which may be computed in closed form by expanding all terms in the Fourier basis. As for the optimization step, it consists in minimizing $g_t(x) - \mu(x)$ over the interval $[0,1]$ which can be done efficiently with exhaustive search.

\begin{figure*}

 \vspace*{-.25cm}

 \includegraphics[scale=.44]{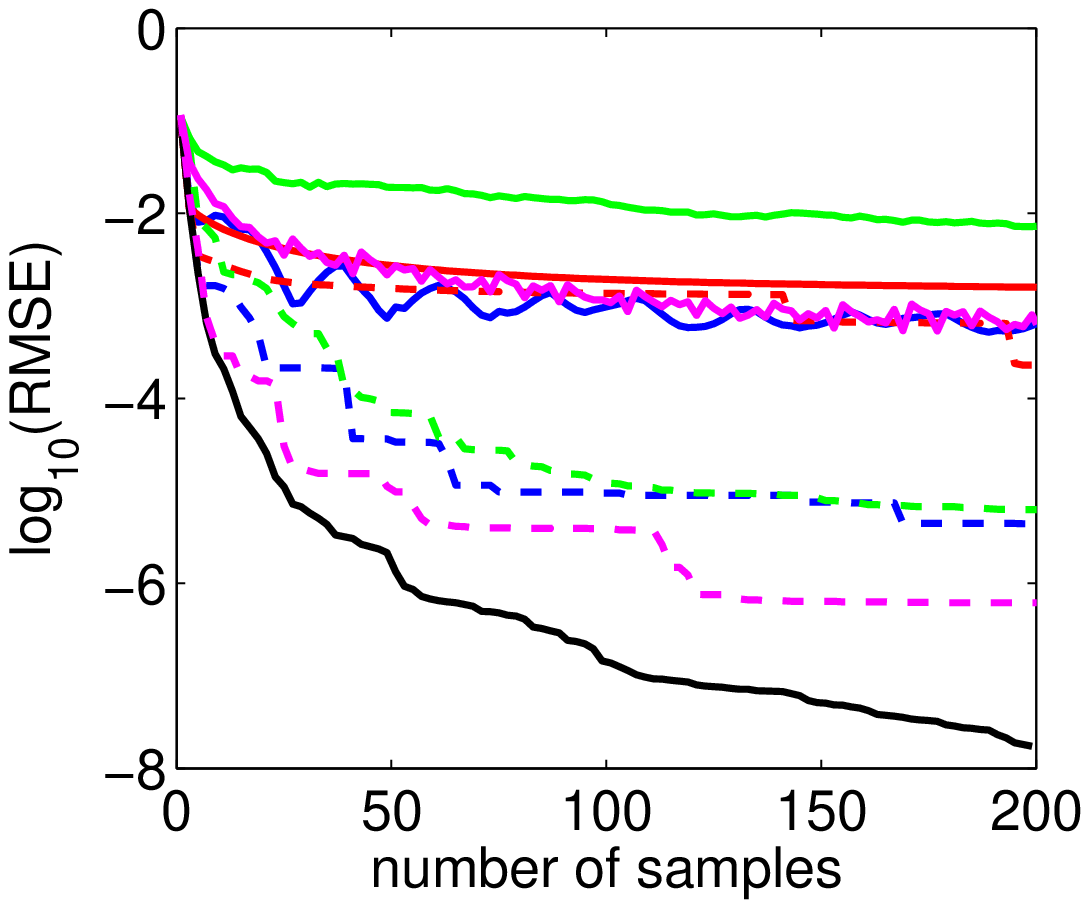} \hspace*{-.2cm}
 \includegraphics[scale=.44]{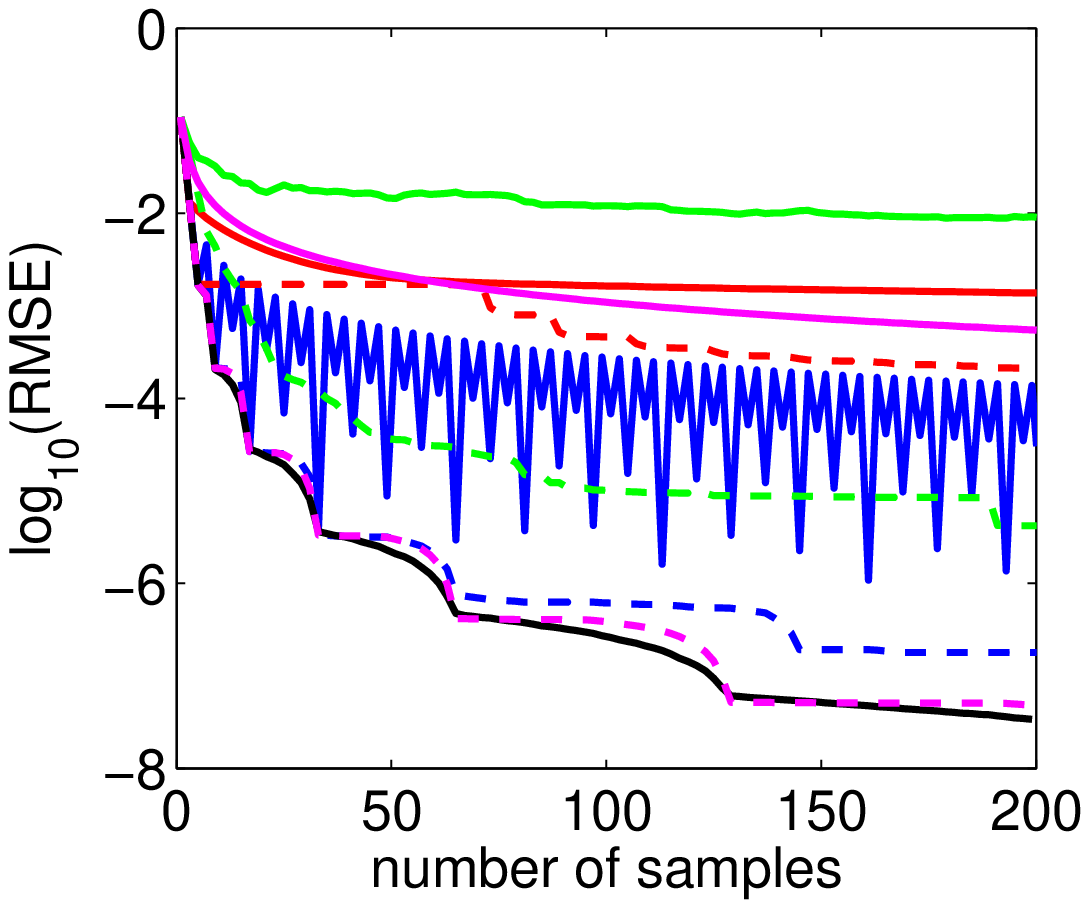} \hspace*{-.2cm}
 \includegraphics[scale=.44]{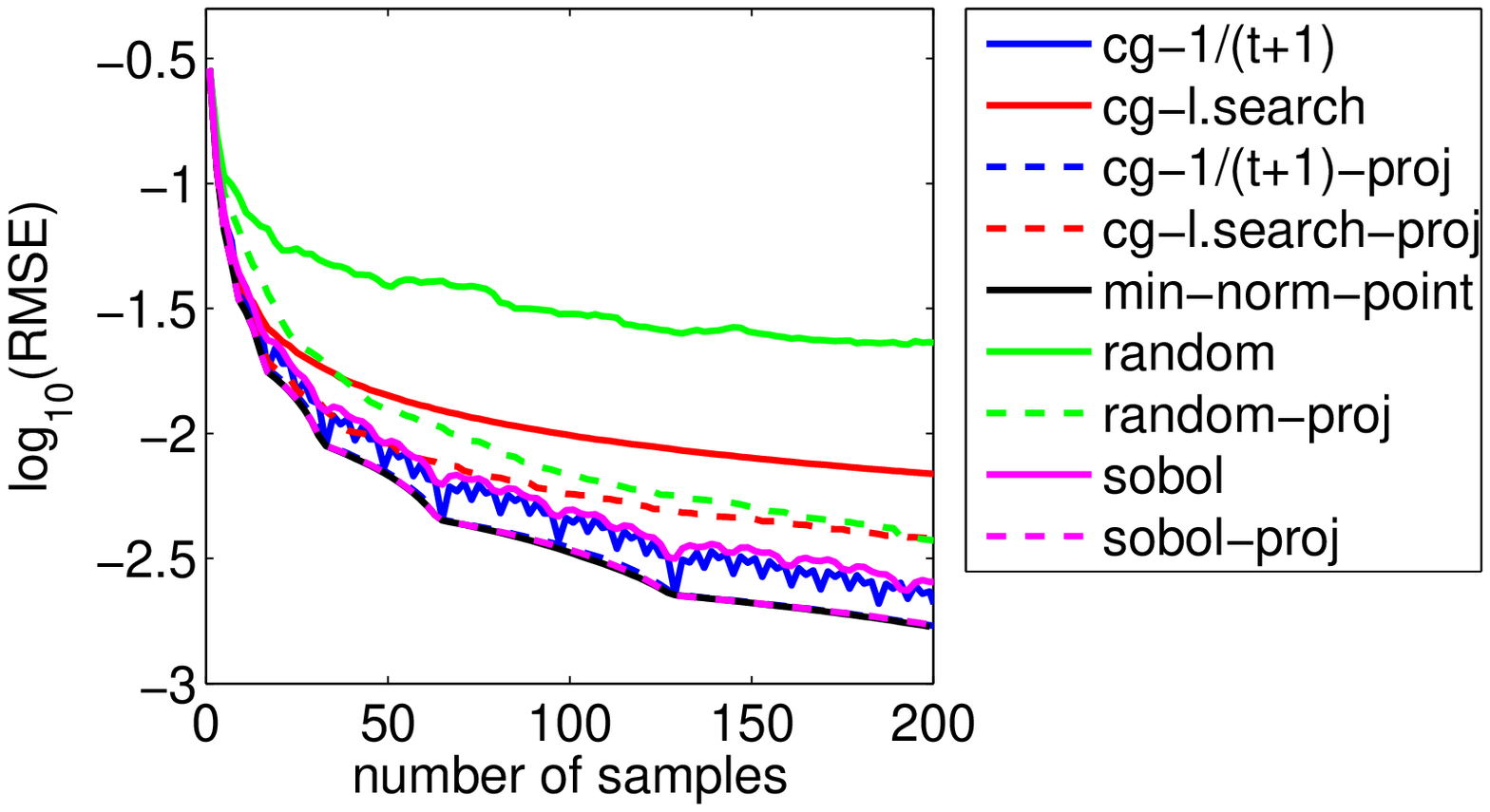}

 \vspace*{-.3cm}

 \caption{Comparison of population herding procedures for kernels on $[0,1]$. From left to right: $\nu=3$ and \emph{non-uniform} density, $\nu=3$ and \emph{uniform} density, $\nu=1$ decay of eigenvalues and \emph{uniform} density. Best seen in color.}
 \label{fig:1}

 \vspace*{-.2cm}

 \end{figure*}

\textbf{Comparison of mean estimation procedures.}
We compare in Figure~\ref{fig:1} two kernels, i.e., with $\nu = 1 $ (left and middle plots) and $\nu =3$ (right plot), the following mean estimation procedures, and plot $\log_{10} \| \hat{\mu} - \mu\|$, for two densities, the uniform density (middle and right) and a randomly selected non-uniform density
(left). We compare the following:

\vspace*{-.4cm}

\begin{list}{\labelitemi}{\leftmargin=1.1em}
   \addtolength{\itemsep}{-.6\baselineskip}
\item[--] \texttt{cg-1/(t+1)}: conditional gradient procedure with   $\rho_t = \frac{1}{t+1}$, which is the original herding procedure of~\citet{welling2009UAIherding}, leading to uniform weights.
\item[--] \texttt{cg-l.search}: conditional gradient procedure with line search (with non-uniform weights).

\item[--] \texttt{min-norm-point}: Minimum-norm point algorithm. This leads to non-uniform weights.
\item[--] \texttt{random}: Random selection of points (from $p(x)$), averaged over 10 replications.
\item[--] \texttt{sobol}: a classical quasi-random sequence, with uniform weights. For non-uniform densities, we first apply the inverse cumulative distribution function.
\end{list}

  \vspace*{-.3cm}

For all of these (except for min-norm-point), we also consider an extra \emph{a posteriori} projection step (denoted by the \texttt{-proj} suffix), which computes the optimal set of non-uniform weights by finding the best approximation of $\mu$ in the convex hull of the points selected by the algorithm.
We can draw the following conclusions:

\vspace*{-.4cm}

\begin{list}{\labelitemi}{\leftmargin=1.1em}
   \addtolength{\itemsep}{-.6\baselineskip}

\item[--] Min-norm-point algorithms always perform best.
\item[--] Conditional gradient with line search is performing slightly worse than regular herding. (Note that we are in the infinite-dimensional setting and so they both have $O(t^{-1})$ as theoretical rate.)

\item[--]  The extra projection step always significantly improves performance, and sometimes enough that random selection of point combined with a reprojection outperforms regular herding (at least for $\nu = 3$, i.e., with a smaller feature space).

\item[--] On the right plot, it turns out that the Sobol sequence is known to achieve the optimal rate of $O(t^{-2})$ for $\|\mu - \hat{\mu}\|^2$ for the associated Sobolev space~\citep{Wah:1990}. In this situation, regular herding empirically achieves the same rate; however, the theoretical analysis provided in the present paper or by~\citet{chensuper} does not allow to explain or support this statement theoretically.

\end{list}

%
%
% \begin{figure}
% \includegraphics[scale=.5]{icml_figure2.eps}
% \caption{Comparison of population herding procedures for kernels on $[0,1]$ with slow decay of eigenvalues and \emph{non-uniform} density.}
% \label{fig:2}
% \end{figure}

\vspace*{-.4cm}

\begin{figure}

 \vspace*{-.25cm}

 \includegraphics[scale=.5]{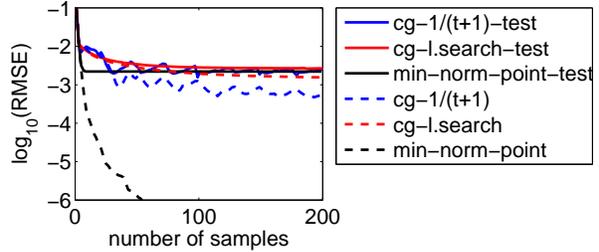}

 \vspace*{-.3cm}

\caption{Comparison of \emph{estimated}  (from a finite sample) herding procedures for kernels on $[0,1]$ with $\nu=3$ and \emph{non-uniform} density.}
 \label{fig:3}

 \vspace*{-.05cm}

 \end{figure}

\textbf{Estimation from a finite sample.} \hspace*{.1cm}
In Figure~\ref{fig:3}, we compare  three of the previously mentioned herding procedures when all quantities are computed from a random sample of size $1000$. In plain, testing errors are computed (using exact expectations) while in dashed, the training errors are computed. All methods eventually fit the empirical mean, with no further progress on the testing error, this behavior happening faster with the min-norm-point algorithm.

% \begin{figure}
% \includegraphics[scale=.5]{icml_figure4.eps}
%\caption{Comparison of \emph{estimated} (from a finite sample) herding procedures for kernels on $[0,1]$ with slow decay of eigenvalues and \emph{non-uniform} density.}
% \label{fig:4}
%
% \end{figure}

\vspace*{-.1cm}

 \subsection{Estimation on graphical models}

\vspace*{-.1cm}

 \label{sec:gm}
 We consider $\mathcal{X} = \{-1,1\}^d$ and a random variable computed as the sign (in $\{-1,1\}$) of a Gaussian random vector in $\rb^d$, together with   $\Phi(x)$ composed of $x$ and of all of its pairwise products $xx^\top$. In this set-up, we can compute the expectation $\E_{p(x)} \Phi(x)$ in closed form, as the mass assigned to the positive orthant by a bivariate Gaussian distribution with correlation $\rho$, which is equal to $\frac{1}{4} + \frac{1}{2\pi} \sin^{-1}\rho$~\citep{abramowitz1964handbook}. We are here in the \emph{finite-dimensional} setting and the faster rates derived in \mysec{rates} apply.

\begin{figure}

 \vspace*{-.35cm}

\centering
 \includegraphics[scale=.4]{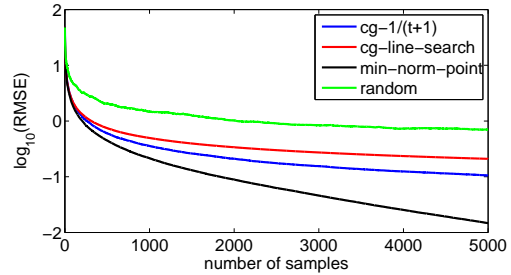}

 \vspace*{-.35cm}

 \caption{Comparison of herding procedures on graphical models with $100$ binary variables. See \mysec{gm}.}
 \label{fig:30}

 \vspace*{-10pt}

 \end{figure}

 We generated $10000$ samples from such a distribution and performed herding with exact expectations but with minima with respect to $x$ computed over this sample (by exhaustive search over the sample). We plot results in \myfig{30}, where we see the superiority of the min-norm-point procedure over the two other  procedures (which include regular herding).  Note that the line-search algorithm is slower than the $1/(t+1)$-rule, which seems to contradict the bounds. The bounds depend on the distance $d$ between the mean and the boundary of the marginal polytope. If this is too small (much like if the strong convexity constant is too small for gradient descent), the linear convergence rate can only be seen for larger numbers of iterations.
%Moreover, in the setting where $d$ is small, the bound for line search depends on $d^2t$ while the bound for the $1/(t+1)$-rule depends on $d t$, thus when $d$ is small, the bound will be smaller for the the $1/(t+1)$-rule for the first few iterations, which we believe happen in this experiment.

% \begin{figure}
% \includegraphics[scale=.5]{icml_figure31.eps}
% \caption{Comparison of herding procedure on graphical models with $100$ binary variables.}
% \label{fig:31}
% \end{figure}

\vspace*{-.1cm}

\subsection{Herding and maximum/minimum entropy}

\vspace*{-.1cm}

Given a moment vector $\mu$ obtained from the empirical mean of $\Phi(x)$ on data, the goal of herding is to produce a pseudo-sample   whose moments match $\mu$ without having to estimate the canonical parameters of the corresponding model. A natural candidate for such a distribution is the maximum entropy distribution and we will compare it to the results of herding in cases where it can be easily computed, namely for $\mathcal{X} = \{-1,1\}^d$ (with $d \leqslant 10$) and either $\Phi(x)=x \in [-1,1]^d$ or $\Phi(x) = (x, x  x^\top)$.
In this setup, following~\citet{welling2009UAIherding}, the distribution on $x \in \mathcal{X}$ is estimated by the empirical distribution $\sum_{i=1}^t w_i \delta(x=x_i)$.

\begin{figure*}

 \vspace*{-.25cm}

 \includegraphics[scale=.445]{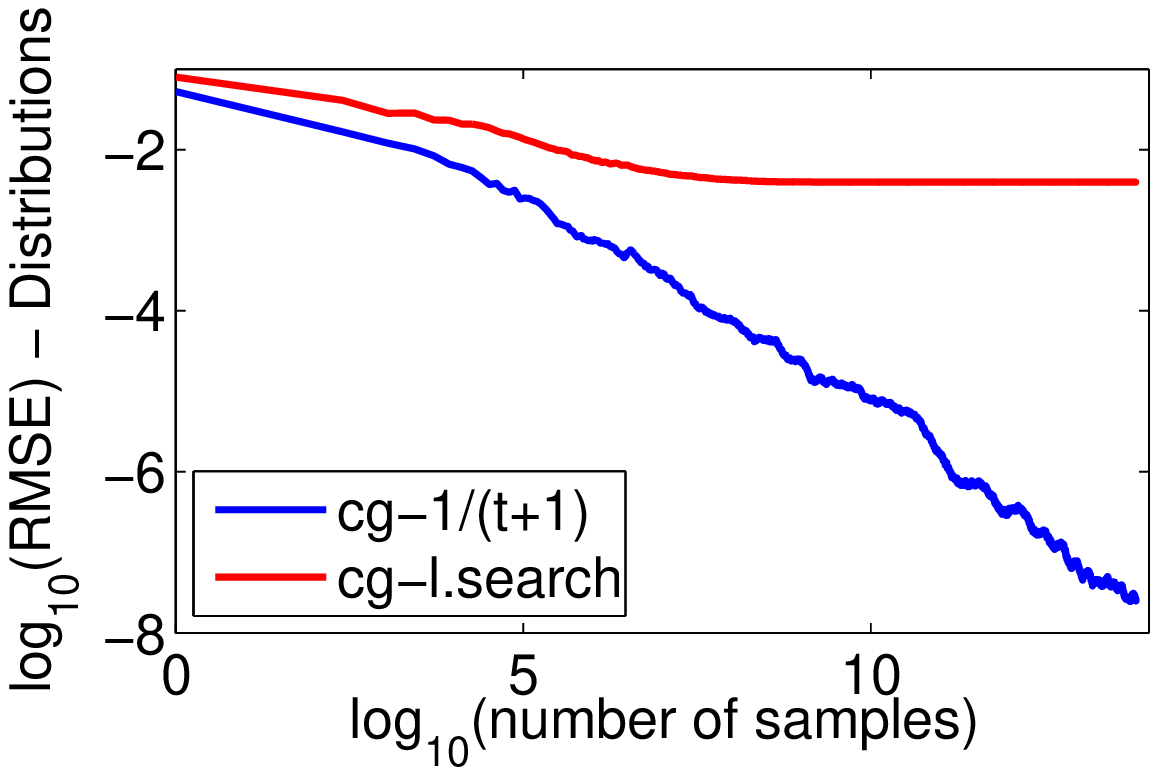} \hspace*{.5cm}  \includegraphics[scale=.445]{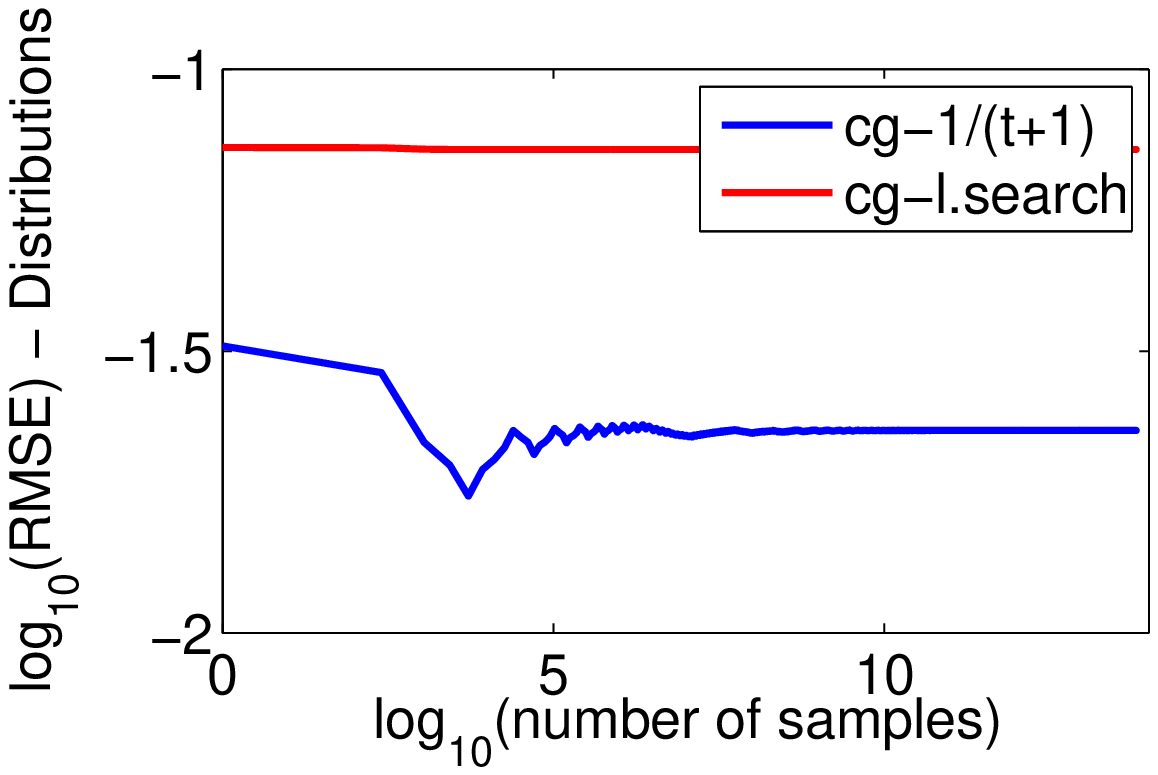}
\hspace*{.5cm}  \includegraphics[scale=.445]{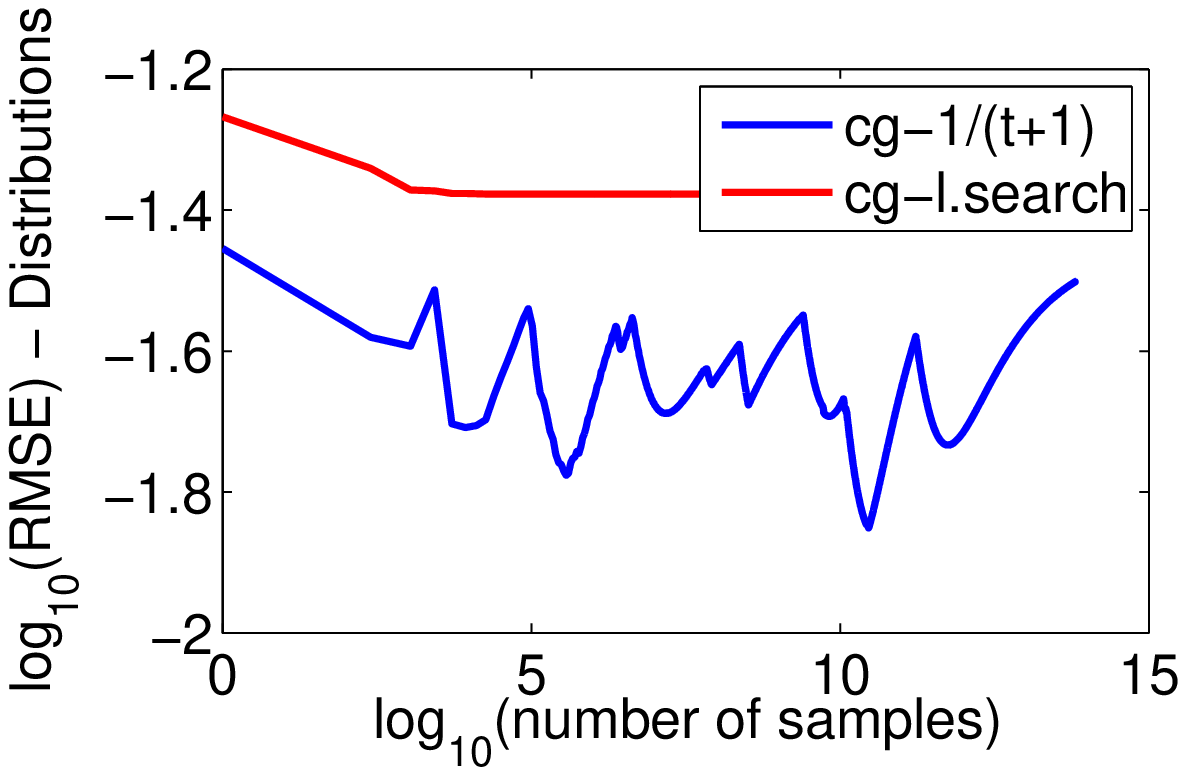}
\\
 \includegraphics[scale=.445]{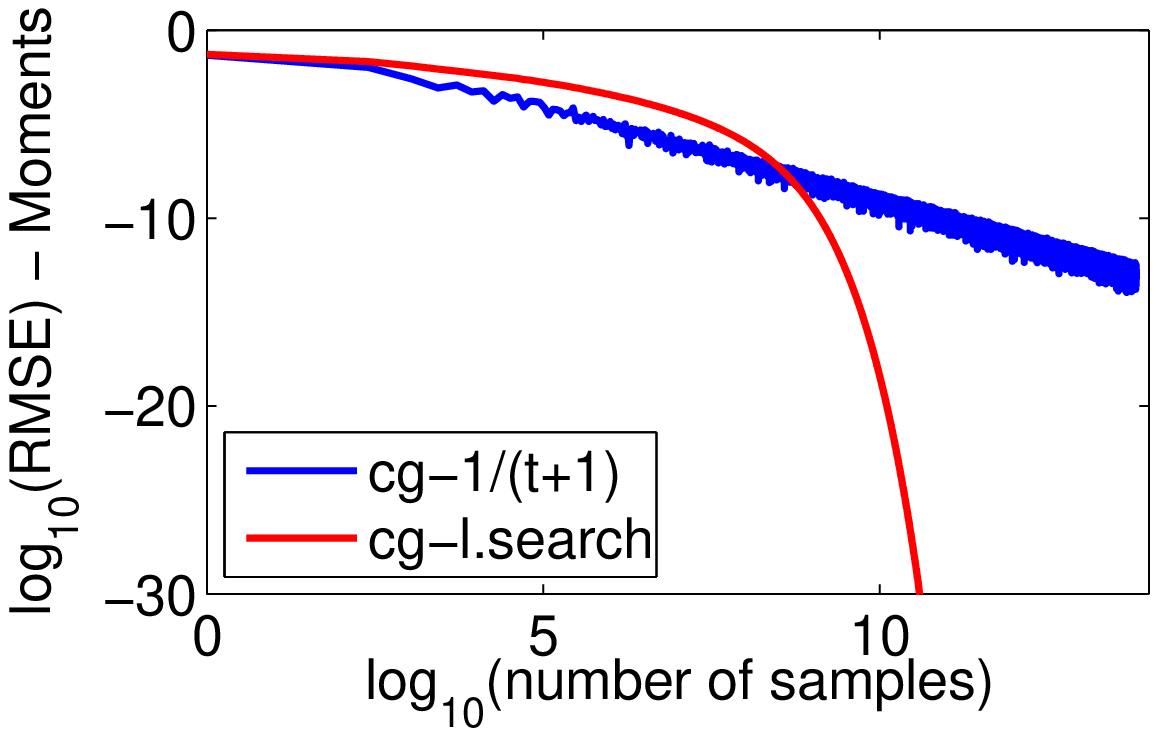} \hspace*{.5cm}  \includegraphics[scale=.445]{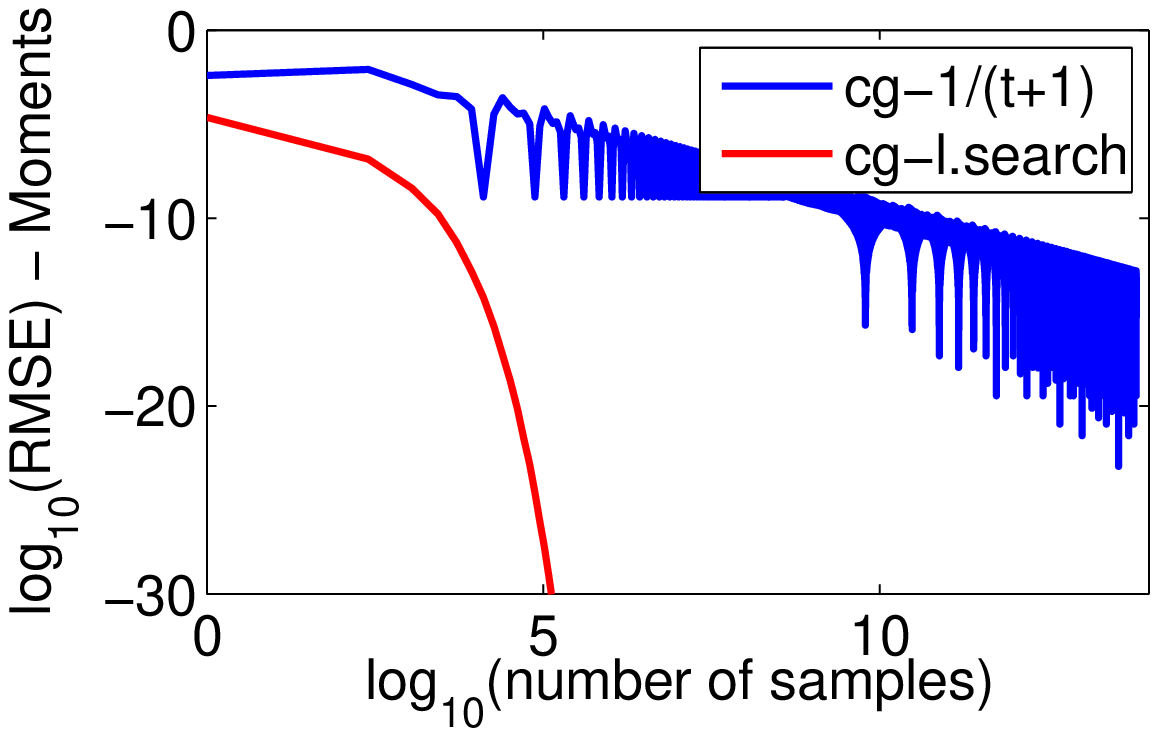}
\hspace*{.5cm}  \includegraphics[scale=.445]{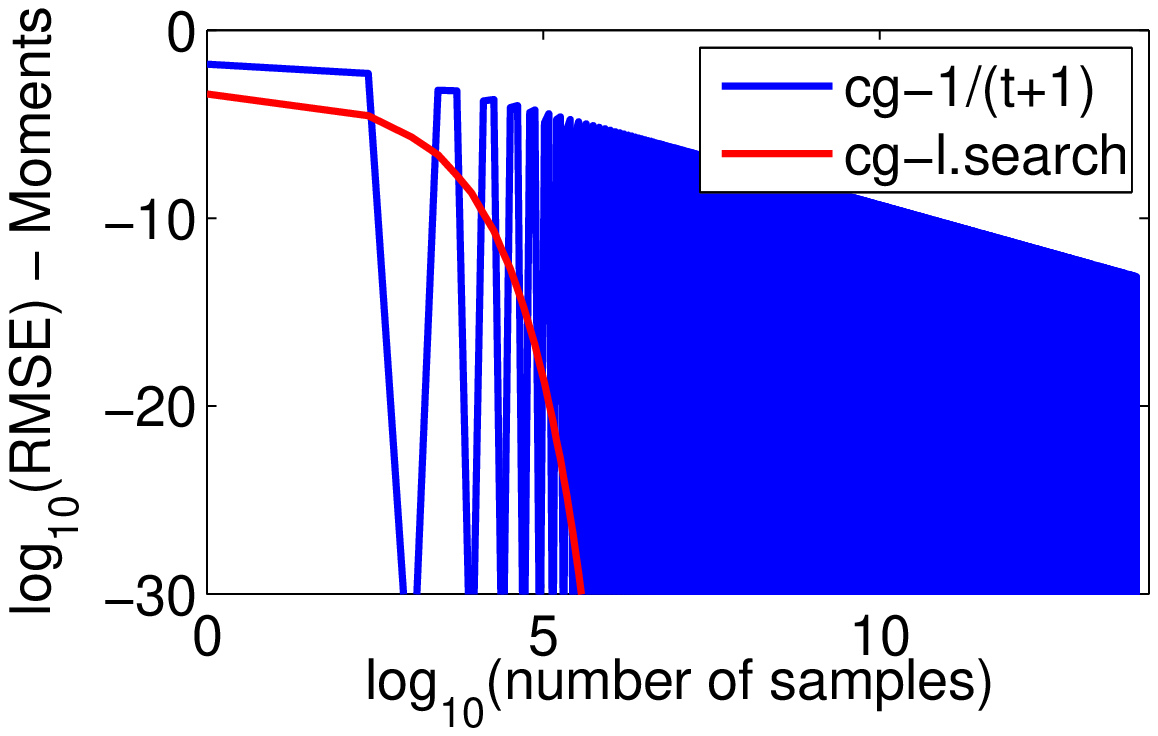}

 \vspace*{-.45cm}

 \caption{Comparison of herding procedures on the independent bit problem with $d=10$ binary variables. Top: estimation of the maximum entropy distribution, bottom: estimation of the mean of the features $\Phi(x)$. From left to right: Mean values are selected uniformly at random on $[-1,1]$, mean values are equal to $\sqrt{2}$ times random rational numbers in  $[-1,1]$, mean values are equal to  random rational numbers in  $[-1,1]$.
 }
 \label{fig:102}

 \vspace*{-.3cm}

 \end{figure*}

\textbf{Learning independent bits.} \hspace*{.05cm}
 \label{sec:bits}
 We first consider $\Phi(x) = x$ and some specific feasible moment $\mu \in \mathcal{M}$. It is well-known that the maximum entropy distribution is the one with independent bits. In the top  panels of Figure~\ref{fig:102}, we compare the norm between the maximum entropy probability vector and the one estimated by two versions of herding, namely conditional gradient with stepsize $\rho_t=1/(t+1)$ (regular herding with uniform weights) and with line search (with non-uniform weights)---the min-norm-point algorithm leads to quantitatively similar results. We show results in \myfig{102} for a mean vector $\mu$ which is a random uniform vector in $[-1,1]^d$ (left plots), and  for a mean  $\mu$ which is random with uniform $(\mu_i+1)/2$ values in $\{1,2,3,4,5\}\times \frac{2 \sqrt{2}}{3}$ (middle plots), and for mean values $\mu$ which is are random with uniform $(\mu_i+1)/2$ values in $\{1,2,3,4,5\}/6 $ (right plots).
 Note that the difference  between rational and irrational means was already brought up by~\citet{welling2010statistical} through the link between herding and  Sturmian sequences.

 For each of the mean vector, we plot in the top plots, the error in estimating the full maximum entropy distribution (a vector of size $2^d$), and in the bottom plots, the error in estimating the feature means (a vector of size $d$). We can draw the following conclusions:

\vspace*{-.4cm}

  \begin{list}{\labelitemi}{\leftmargin=1.1em}
   \addtolength{\itemsep}{-.6\baselineskip}

 \item[--] For a random vector $\mu$ (left plots), then regular herding (with no line search) empirically converges to the maximum entropy distribution.
 \item[--] For rational ratios between the means (but irrational means, middle plots), then there is no convergence to the maximum entropy distribution.
 \item[--] For rational means (right), there is no convergence either, but the behavior is more erratic.
 \item[--] The line-search procedure never converges to the maximum entropy procedure. On the opposite, it happens to be close to a minimum entropy solution, where many states have probability zero.
 \end{list}

 \vspace*{-.4cm}

 Experiments considered in \myfig{102} considered a single draw of the mean vector $\mu$, but similar empirical conclusions may be drawn from other random samples, and we conjecture that for \emph{almost surely} all random vectors $\mu \in [-1,1]^d$ (which would avoid rational ratios between mean values), then regular herding converges to the maximum entropy distribution. The next experiment shows that this is not the case in general.

\begin{figure}

 \vspace*{-.25cm}

\centering
 \includegraphics[scale=.445]{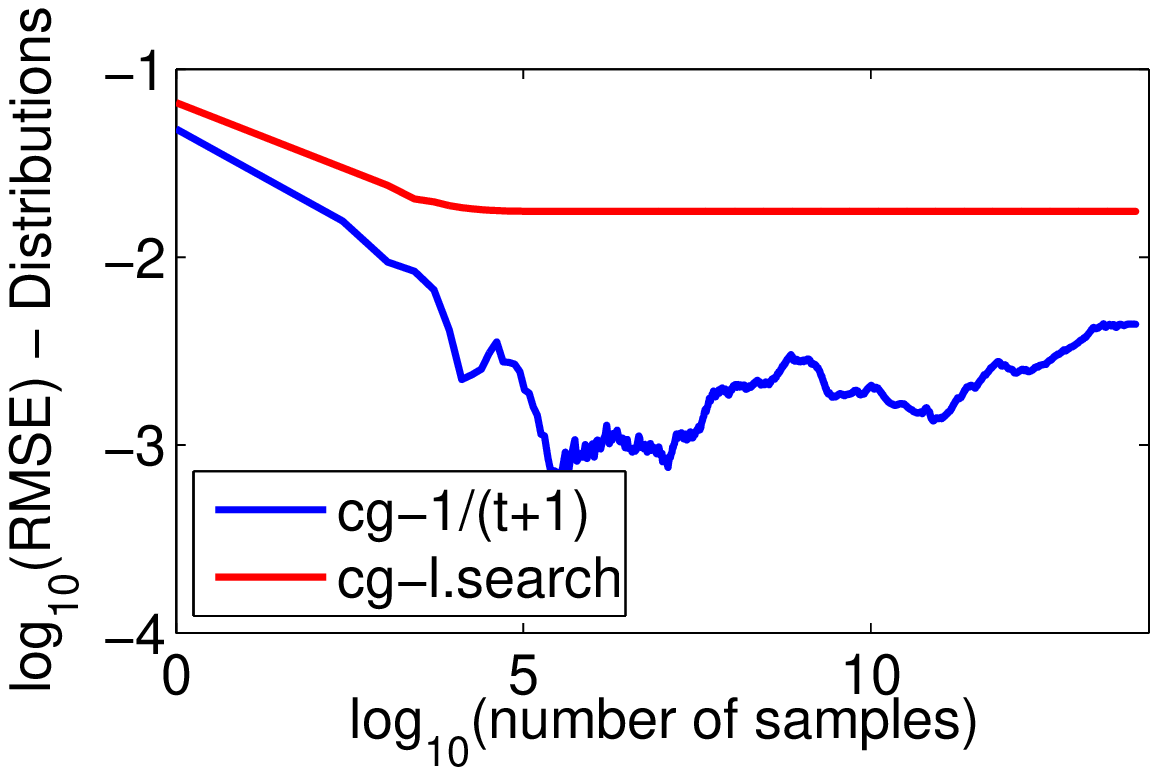}
\\
 \includegraphics[scale=.445]{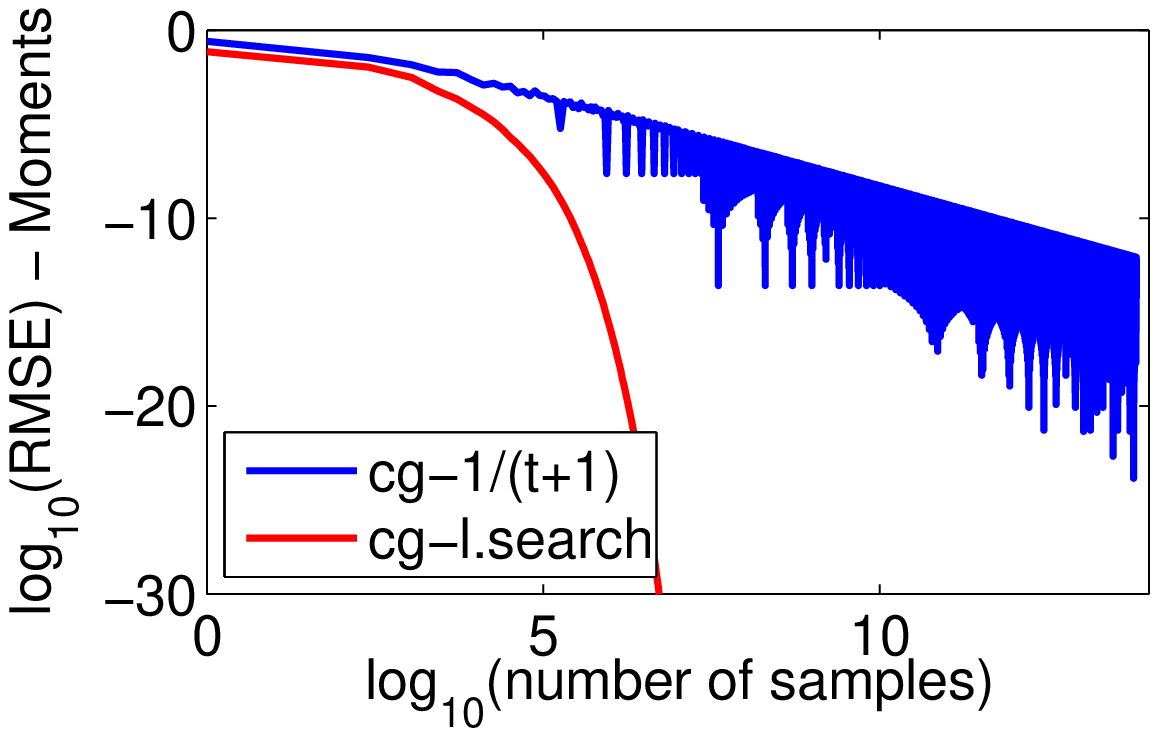}

 \vspace*{-.45cm}

 \caption{Comparison of herding procedures on graphical models with $10$ binary variables.  Top: estimation of the maximum entropy distribution, bottom: estimation of the mean of the features $\Phi(x)$.}
 \label{fig:104}

 \vspace*{-.5cm}

 \end{figure}

\textbf{Learning non-independent bits.} \hspace*{.05cm}
 We now consider $\Phi(x) = (x,x x^\top)$, and a certain random feasible moment $\mu \in \mathcal{M}$. As before, we compare the norm between the maximum entropy probability vector and the one estimated by the two versions of herding. We present results in \myfig{104} for a mean vector obtained by the corresponding exponential family distribution with zero weights for the features $x$ and constant weights on the features $xx^\top$. We see that the herding procedures, while leading to a consistent estimation of the mean vector, does not converge to the maximum entropy distribution and other unreported experiments have led to similar results.

%\vspace*{-5pt}

\vspace*{-.1cm}

\section{Conclusion}

\vspace*{-.1cm}

We showed that herding generates a sequence of points which give in sequence the
descent directions of a conditional gradient algorithm
minimizing the quadratic error on the moment vector. Therefore,
if herding is only assessed in terms of its ability to approximate the
moment vector, it is outperformed by other more efficient algorithms.
Clearly, herding was originally defined with another goal, which was
to generate a pseudo-sample whose distribution could approach the
maximum entropy distribution with a given moment vector. Our
experiments suggest empirically, that while this is the case in
certain cases, herding fails in other case, which are not chosen to be
particularly pathological. This probably prompts for a
further study of herding.

\vspace*{-.1cm}

Our experiments also show that
algorithms that are more efficient than herding at approximating the
moment vector fail more blatantly to approach a maximum entropy
distribution and they present characteristics which would rather
suggest a minimization of the entropy.
This suggests the question of whether there is a tradeoff between
approximating most efficiently the mean vector and
approximating well the maximum entropy distribution, or if the two
goals are in fact rather aligned?
In any case, we hope that formulating herding as an optimization
problem can help form a better understanding of its goals and its properties.

\vspace*{-.1cm}

\textbf{Acknowledgements.} \hspace*{.1cm}
We thank Ferenc Husz\'{a}r and Zoubin Ghahramani for helpful discussions. This work was supported by  the European Research
Council (SIERRA Project) and the city of Paris (``Research in Paris'' program).

\vspace*{-.3cm}

\small{\bibliography{herding}}

\begin{thebibliography}{17}
\providecommand{\natexlab}[1]{#1}
\providecommand{\url}[1]{\texttt{#1}}
\expandafter\ifx\csname urlstyle\endcsname\relax
  \providecommand{\doi}[1]{doi: #1}\else
  \providecommand{\doi}{doi: \begingroup \urlstyle{rm}\Url}\fi

\bibitem[Abramowitz \& Stegun(1964)Abramowitz and
  Stegun]{abramowitz1964handbook}
Abramowitz, M. and Stegun, I.A.
\newblock \emph{Handbook of mathematical functions}.
\newblock Dover publications, 1964.

\bibitem[Bach(2011)]{bach2011learning}
Bach, F.
\newblock Learning with submodular functions: A convex optimization
  perspective.
\newblock Technical Report 1111.6453, Arxiv, 2011.

\bibitem[Beck \& Teboulle(2004)Beck and Teboulle]{beck2004conditional}
Beck, A. and Teboulle, M.
\newblock A conditional gradient method with linear rate of convergence for
  solving convex linear systems.
\newblock \emph{Math. Meth. Op. Res.}, 59\penalty0 (2):\penalty0 235--247,
  2004.

\bibitem[Boucheron et~al.(2005)Boucheron, Bousquet, and
  Lugosi]{boucheron2005theory}
Boucheron, S., Bousquet, O., and Lugosi, G.
\newblock Theory of classification: A survey of some recent advances.
\newblock \emph{ESAIM Probability and statistics}, 9:\penalty0 323--375, 2005.

\bibitem[Chen et~al.(2010)Chen, Welling, and Smola]{chensuper}
Chen, Y., Welling, M., and Smola, A.
\newblock Super-samples from kernel herding.
\newblock In \emph{Proc. UAI}, 2010.

\bibitem[Cucker \& Smale(2002)Cucker and Smale]{smale}
Cucker, F. and Smale, S.
\newblock On the mathematical foundations of learning.
\newblock \emph{Bull. AMS}, 39\penalty0 (1), 2002.

\bibitem[Dunn(1980)]{dunn1980convergence}
Dunn, J.~C.
\newblock Convergence rates for conditional gradient sequences generated by
  implicit step length rules.
\newblock \emph{SIAM J. Control \& Opt.}, 18:\penalty0 473--487, 1980.

\bibitem[Gelfand et~al.(2010)Gelfand, van~der Maaten, Chen, and
  Welling]{gelfand2010herding}
Gelfand, A., van~der Maaten, L., Chen, Y., and Welling, M.
\newblock On herding and the perceptron cycling theorem.
\newblock In \emph{Adv. NIPS}, 2010.

\bibitem[Morokoff \& Caflisch(1994)Morokoff and Caflisch]{morokoff1994quasi}
Morokoff, W.J. and Caflisch, R.E.
\newblock Quasi-random sequences and their discrepancies.
\newblock \emph{SIAM Journal on Scientific Computing}, 15\penalty0
  (6):\penalty0 1251--1279, 1994.

\bibitem[O'Hagan(1991)]{o1991bayes}
O'Hagan, A.
\newblock Bayes-{H}ermite quadrature.
\newblock \emph{J. Stat. Planning \& Inference}, 29\penalty0 (3):\penalty0
  245--260, 1991.

\bibitem[Smola et~al.(2007)Smola, Gretton, Song, and
  Sch{\"o}lkopf]{smola2007hilbert}
Smola, A., Gretton, A., Song, L., and Sch{\"o}lkopf, B.
\newblock A {H}ilbert space embedding for distributions.
\newblock In \emph{Algorithmic Learning Theory}, pp.\  13--31. Springer, 2007.

\bibitem[Wahba(1990)]{Wah:1990}
Wahba, Grace.
\newblock \emph{Spline Models for Observational Data}.
\newblock SIAM, 1990.

\bibitem[Wainwright \& Jordan(2008)Wainwright and
  Jordan]{wainwright2008graphical}
Wainwright, M.J. and Jordan, M.I.
\newblock Graphical models, exponential families, and variational inference.
\newblock \emph{Found. Trends Mach. Learn.}, 1\penalty0 (1-2):\penalty0 1--305,
  2008.

\bibitem[Welling(2009{\natexlab{a}})]{welling2009UAIherding}
Welling, M.
\newblock Herding dynamic weights for partially observed random field models.
\newblock In \emph{Proc. UAI}, 2009{\natexlab{a}}.

\bibitem[Welling(2009{\natexlab{b}})]{welling2009herding}
Welling, M.
\newblock Herding dynamical weights to learn.
\newblock In \emph{Proc. ICML}, 2009{\natexlab{b}}.

\bibitem[Welling \& Chen(2010)Welling and Chen]{welling2010statistical}
Welling, M. and Chen, Y.
\newblock Statistical inference using weak chaos and infinite memory.
\newblock \emph{J. Physics: Conf. Series}, 233, 2010.

\bibitem[Wolfe(1976)]{wolfe1976finding}
Wolfe, P.
\newblock Finding the nearest point in a polytope.
\newblock \emph{Math. Progr.}, 11\penalty0 (1):\penalty0 128--149, 1976.

\end{thebibliography}
\bibliographystyle{icml2012}

\end{document}